\documentclass{article}
\usepackage{arxiv}

\usepackage[T1]{fontenc}    
\usepackage{hyperref}       
\usepackage{url}            
\usepackage{booktabs}       
\usepackage{amsfonts}       
\usepackage{nicefrac}       
\usepackage{microtype}      
\usepackage{lipsum}		
\usepackage{graphicx}
\usepackage{natbib}
\usepackage{doi}
\usepackage{pythontex}

\title{Generalized Linear Tree Space Nearest Neighbor}

\author{ Michael S Kim  \\
	Gradient Ascent LLC \\
	\texttt{mikeskim @ gmail . com} \\

}

\renewcommand{\shorttitle}{Michael S Kim}

\hypersetup{
pdftitle={},
pdfsubject={cs.LG},
pdfauthor={Michael S Kim},
pdfkeywords={},
}

\begin{document}
\maketitle

\begin{abstract}
We present a novel method of stacking decision trees by projection into an ordered time split out-of-fold (OOF) one nearest neighbor (1NN) space. The predictions of these one nearest neighbors are combined through a linear model. This process is repeated many times and averaged to reduce variance. Generalized Linear Tree Space Nearest Neighbor (GLTSNN) is competitive with respect to Mean Squared Error (MSE) compared to Random Forest (RF) on several publicly available datasets. Some of the theoretical and applied advantages of GLTSNN are discussed. We conjecture a classifier based upon the GLTSNN would have an error that is asymptotically bounded by twice the Bayes error rate like k = 1 Nearest Neighbor. 
\end{abstract}



\section{Introduction}
Stacked Generalization \citep{WOLPERT1992241} is often used in competitive machine learning to increase model predictive power on held out data. However, the typical methods of stacking result in explicit target leakage \citep{DBLP:journals/corr/DorogushGGKPV17}. To avoid all explicit target leakage, it is possible to create an artificial psuedo-random time index for each data point. Then the arrow of time is respected such that only prior information is used in training an estimator. The trained estimator can then predict on future data. These leak free predictions can then be used as inputs to other estimators that also respect the arrow of time. In particular, we utilize Cascade Generalization in our ensemble estimator \citep{10.1023/A:1007652114878} to lower bias. Then we average across many different 1NN meta feature based linear models to reduce variance. This final step of our estimator leaks the target as the arrow of time is not respected and all data is used. However, such leakage is the price paid for increased estimator power. We show this new estimator with reduced bias and reduced variance is highly competitive with Random Forest (RF) \cite{breiman2001random}.

\section{Methods} 
We implement our novel algorithm in Python (3.8.5) \citep{van1995python} with Pandas (1.1.5) \citep{reback2020pandas} \citep{mckinney-proc-scipy-2010}, Numpy (1.19.4) \citep{harris2020array}, and Sklearn (0.23.2) \citep{scikit-learn}. At a high level, our algorithm fits decision tree (extra tree) estimators on time based splits of the input data. These trees form a cascade where the predictions of prior trees are fed into future trees. The second stage after the cascade of trees consists of many 1NN estimators on different shuffles of the input data where the feature space is the out-of-fold (OOF) tree space. Notice that in general these 1NN when used on training data does not output the data point itself. The final meta estimator takes the output of many 1NN estimators as input to a linear model (Bayesian Ridge) and outputs a final prediction. 

While this estimator can be implemented in any language, the exact Python code used in all our experiments is presented here. 
\begin{pyblock}
import copy
import pandas as pd
import numpy as np
from sklearn.linear_model import BayesianRidge
from sklearn.neighbors import KNeighborsRegressor
from sklearn.base import BaseEstimator, RegressorMixin
from sklearn.tree import ExtraTreeRegressor
\end{pyblock}

\begin{pyblock}
class LTNNRegressor(BaseEstimator, RegressorMixin):
    def __init__(self, random_seed=0, tree_depth=None, num_folds=10, num_knn=100):
        self.random_seed = random_seed
        self.tree_depth = tree_depth
        self.num_folds = num_folds
        self.num_knn = num_knn
        self.tree_list = []
        self.knn_list = []
        self.final_model = None

    def fit(self, X, y):
        X = pd.DataFrame(X, copy=True)
        features_list = X.columns.tolist()
        X["___target"] = y
        knn = KNeighborsRegressor(n_neighbors=1, n_jobs=-1)
        dt = ExtraTreeRegressor(
            max_depth=self.tree_depth, random_state=self.random_seed
        )
        linear = BayesianRidge(normalize=True)
        knn_list = []
        final_features = []
        for n_seed in range(self.num_knn):
            X = X.sample(frac=1, random_state=n_seed + self.random_seed).reset_index(
                drop=True
            )
            X["___index"] = range(len(X))
            knn_features = []
            tree_list = []
            for n_fold in range(self.num_folds - 1):
                train_index = X["___index"] < ((n_fold + 1.0) / self.num_folds) * len(X)
                valid_index = ~train_index
                X_train = X[train_index]
                X_valid = X[valid_index]
                dt.fit(X_train[features_list + knn_features], X_train["___target"])
                tree_list.append(copy.deepcopy(dt))
                preds = dt.predict(X_valid[features_list + knn_features])
                X["___knn_" + str(n_fold)] = np.nan
                X.loc[valid_index, "___knn_" + str(n_fold)] = preds
                X["___knn_" + str(n_fold)].fillna(
                    X["___knn_" + str(n_fold)].mean(), inplace=True
                )
                knn_features.append("___knn_" + str(n_fold))
            self.tree_list.append(copy.deepcopy(tree_list))
            knn.fit(X[knn_features], X["___target"])
            knn_list.append(copy.deepcopy(knn))
            X["___final_" + str(n_seed)] = knn.predict(X[knn_features])
            final_features.append("___final_" + str(n_seed))
        self.knn_list = copy.deepcopy(knn_list)
        linear.fit(X[final_features], X["___target"])
        self.final_model = copy.deepcopy(linear)
        return self

    def predict(self, X):
        X_copy = pd.DataFrame(X, copy=True)
        final_preds = []
        for num, knn in enumerate(self.knn_list):
            X = X_copy.copy()
            X_meta = pd.DataFrame()
            for feat, tree in enumerate(self.tree_list[num]):
                X_meta["___knn_" + str(feat)] = tree.predict(X)
                X["___knn_" + str(feat)] = X_meta["___knn_" + str(feat)].values
            final_preds += [knn.predict(X_meta)]
        X_final = np.column_stack(final_preds)
        final_preds = self.final_model.predict(X_final)
        return final_preds
\end{pyblock}
We note some interesting applied properties of this estimator. First, the difficulty of normalizing data and weighting features for 1NN is not present in our algorithm. This is because the cascade of trees does this projection for us. This means the estimator can be used like Random Forest on even categorical input features. Second, the 1NN does not overfit to the degree you would expect 1NN to overfit because the prior level of estimators respect the arrow of time when making OOF predictions. Third, many rows for many features that are input into the 1NN are imputed with the OOF decision tree prediction mean which causes very slight target leakage. This can be taken out by using imputation by 0 or some other constant. However, this does not work as well as mean based imputation given the final model is linear in nature. Lastly, any linear model can be used as the final stage estimator. In our implementation we use Bayesian Ridge Regression, but we could have easily used a Generalized Linear Model with a Gamma distribution or Robust Regression with a Huber loss function. This allows flexibility with respect to loss functions depending on the type of problem that needs to be solved.

\section{Experiments}
We conduct three experiments on three different datasets. All three datasets are regression problems where the target is real valued. The datasets include:
\begin{itemize}
	\item Boston house-prices
	\item Diabetes disease progression
	\item Friedman 1 simulation
\end{itemize}
We use the default parameters for sklearn's Random Forest. The parameters of our novel estimator GLTSNN is also fixed across all three datasets. The fixed default parameters are exactly as delineated in our code. Five fold cross validation is used with a fixed seed for both estimators. The models are evaluated on OOF MSE. 
\begin{pyblock}
def run_experiments(reg):
    from sklearn.datasets import load_diabetes, load_boston, make_friedman1
    from sklearn.metrics import mean_squared_error
    from sklearn.model_selection import cross_val_predict, KFold
    
    oof_mse = []
    data_list = [load_diabetes(return_X_y=True), 
                 load_boston(return_X_y=True),
                 make_friedman1(n_samples=1000, random_state=0)]
    for X,y in data_list:
        kf = KFold(n_splits=5, shuffle=True, random_state=0)
        y_pred = cross_val_predict(reg, X, y, cv=kf)
        mse_val = mean_squared_error(y, y_pred)
        oof_mse.append(mse_val)
    return oof_mse
\end{pyblock}

\section{Results}
The MSE of our novel estimator is lower than Random Forest for all three experiments. We do not imply that our estimator is better than Random Forest in general, or even on these datasets. It is possible a highly tuned version of Random Forest or Extra Trees would beat GLTSNN. However, these results show our estimator is competitive with other ensemble methods such as RF \ref{tab:table}.

\begin{table}
	\caption{OOF MSE: RF vs. GLTSNN}
	\centering
	\begin{tabular}{lll}
		\toprule
		\multicolumn{2}{c}{Experiments}                   \\
		\cmidrule(r){1-2}
		Estimator     & Dataset     & MSE  \\
		\midrule
		RF & Boston  & 14.82    \\
		RF     & Diabetes & 3408.61    \\
		RF     & Friedman       & 3.09  \\
		GLTSNN & Boston  & 12.21   \\
		GLTSNN     & Diabetes & 3253.13    \\
	  GLTSNN     & Friedman       &  2.60 \\
		\bottomrule
	\end{tabular}
	\label{tab:table}
\end{table}

\section{Conclusion}
We have constructed a novel estimator (regressor) based upon stacking, cascading, and pseudo time based fold generation. The foundations of the estimator suggest a reduction in both bias and variance. Furthermore we conjecture this estimator to have lower correlation with other tree based methods because of it utilizes a 1NN based meta estimator. Future research may consider extending this to estimator to classification. A open theoretical question is whether or not this estimator has an error rate that is asymptotically bounded by a constant less than or equal to two times the Bayes error rate. Future applied work could focus upon time efficiency through approximate nearest neighbor, and online estimation methods.

\bibliographystyle{unsrtnat}
\bibliography{references}  

\begin{thebibliography}{9}
\providecommand{\natexlab}[1]{#1}
\providecommand{\url}[1]{\texttt{#1}}
\expandafter\ifx\csname urlstyle\endcsname\relax
  \providecommand{\doi}[1]{doi: #1}\else
  \providecommand{\doi}{doi: \begingroup \urlstyle{rm}\Url}\fi

\bibitem[Wolpert(1992)]{WOLPERT1992241}
David~H. Wolpert.
\newblock Stacked generalization.
\newblock \emph{Neural Networks}, 5\penalty0 (2):\penalty0 241--259, 1992.
\newblock ISSN 0893-6080.
\newblock \doi{https://doi.org/10.1016/S0893-6080(05)80023-1}.
\newblock URL
  \url{https://www.sciencedirect.com/science/article/pii/S0893608005800231}.

\bibitem[Dorogush et~al.(2017)Dorogush, Gulin, Gusev, Kazeev, Prokhorenkova,
  and Vorobev]{DBLP:journals/corr/DorogushGGKPV17}
Anna~Veronika Dorogush, Andrey Gulin, Gleb Gusev, Nikita Kazeev,
  Liudmila~Ostroumova Prokhorenkova, and Aleksandr Vorobev.
\newblock Fighting biases with dynamic boosting.
\newblock \emph{CoRR}, abs/1706.09516, 2017.
\newblock URL \url{http://arxiv.org/abs/1706.09516}.

\bibitem[Gama and Brazdil(2000)]{10.1023/A:1007652114878}
Jo\~{a}o Gama and Pavel Brazdil.
\newblock Cascade generalization.
\newblock \emph{Mach. Learn.}, 41\penalty0 (3):\penalty0 315–343, December
  2000.
\newblock ISSN 0885-6125.
\newblock \doi{10.1023/A:1007652114878}.
\newblock URL \url{https://doi.org/10.1023/A:1007652114878}.

\bibitem[Breiman(2001)]{breiman2001random}
Leo Breiman.
\newblock Random forests.
\newblock \emph{Machine Learning}, 45\penalty0 (1):\penalty0 5--32, 2001.
\newblock ISSN 0885-6125.
\newblock \doi{10.1023/A:1010933404324}.
\newblock URL \url{http://dx.doi.org/10.1023/A%3A1010933404324}.

\bibitem[Van~Rossum and Drake~Jr(1995)]{van1995python}
Guido Van~Rossum and Fred~L Drake~Jr.
\newblock \emph{Python tutorial}.
\newblock Centrum voor Wiskunde en Informatica Amsterdam, The Netherlands,
  1995.

\bibitem[pandas~development team(2020)]{reback2020pandas}
The pandas~development team.
\newblock pandas-dev/pandas: Pandas, February 2020.
\newblock URL \url{https://doi.org/10.5281/zenodo.3509134}.

\bibitem[{W}es {M}c{K}inney(2010)]{mckinney-proc-scipy-2010}
{W}es {M}c{K}inney.
\newblock {D}ata {S}tructures for {S}tatistical {C}omputing in {P}ython.
\newblock In {S}t\'efan van~der {W}alt and {J}arrod {M}illman, editors,
  \emph{{P}roceedings of the 9th {P}ython in {S}cience {C}onference}, pages 56
  -- 61, 2010.
\newblock \doi{10.25080/Majora-92bf1922-00a}.

\bibitem[Harris et~al.(2020)Harris, Millman, van~der Walt, Gommers, Virtanen,
  Cournapeau, Wieser, Taylor, Berg, Smith, Kern, Picus, Hoyer, van Kerkwijk,
  Brett, Haldane, del R{'{\i}}o, Wiebe, Peterson, G{'{e}}rard-Marchant,
  Sheppard, Reddy, Weckesser, Abbasi, Gohlke, and Oliphant]{harris2020array}
Charles~R. Harris, K.~Jarrod Millman, St{'{e}}fan~J. van~der Walt, Ralf
  Gommers, Pauli Virtanen, David Cournapeau, Eric Wieser, Julian Taylor,
  Sebastian Berg, Nathaniel~J. Smith, Robert Kern, Matti Picus, Stephan Hoyer,
  Marten~H. van Kerkwijk, Matthew Brett, Allan Haldane, Jaime~Fern{'{a}}ndez
  del R{'{\i}}o, Mark Wiebe, Pearu Peterson, Pierre G{'{e}}rard-Marchant, Kevin
  Sheppard, Tyler Reddy, Warren Weckesser, Hameer Abbasi, Christoph Gohlke, and
  Travis~E. Oliphant.
\newblock Array programming with {NumPy}.
\newblock \emph{Nature}, 585\penalty0 (7825):\penalty0 357--362, September
  2020.
\newblock \doi{10.1038/s41586-020-2649-2}.
\newblock URL \url{https://doi.org/10.1038/s41586-020-2649-2}.

\bibitem[Pedregosa et~al.(2011)Pedregosa, Varoquaux, Gramfort, Michel, Thirion,
  Grisel, Blondel, Prettenhofer, Weiss, Dubourg, Vanderplas, Passos,
  Cournapeau, Brucher, Perrot, and Duchesnay]{scikit-learn}
F.~Pedregosa, G.~Varoquaux, A.~Gramfort, V.~Michel, B.~Thirion, O.~Grisel,
  M.~Blondel, P.~Prettenhofer, R.~Weiss, V.~Dubourg, J.~Vanderplas, A.~Passos,
  D.~Cournapeau, M.~Brucher, M.~Perrot, and E.~Duchesnay.
\newblock Scikit-learn: Machine learning in {P}ython.
\newblock \emph{Journal of Machine Learning Research}, 12:\penalty0 2825--2830,
  2011.

\end{thebibliography}

\end{document}